\documentclass[11pt]{article}
\usepackage{times}
\usepackage{url}
\usepackage{latexsym}

\usepackage{hyperref}
\usepackage{url}
\usepackage{amsmath}
\usepackage{centernot}
\usepackage{graphicx,array,multirow}
\usepackage{eacl2017}

\eaclfinalcopy %

\title{Using the Output Embedding to Improve Language Models}

\author{Ofir Press \and   Lior Wolf \\
  School of Computer Science\\
  Tel-Aviv University, Israel \\
  {\tt \{ofir.press,wolf\}@cs.tau.ac.il} 
}

\date{}

\begin{document}

\setlength{\abovedisplayskip}{3pt}
\setlength{\belowdisplayskip}{3pt}

\maketitle
\begin{abstract}
We study the topmost weight matrix of neural network language models. We show that this matrix constitutes a valid word embedding. When training language models, we recommend tying the input embedding and this output embedding. We analyze the resulting update rules and show that the tied embedding evolves in a more similar way to the output embedding than to the input embedding in the untied model. We also offer a new method of regularizing the output embedding. Our methods lead to a significant reduction in perplexity, as we are able to show on a variety of neural network language models. Finally, we show that weight tying can reduce the size of neural translation models to less than half of their original size without harming their performance. 
\end{abstract}

\section{Introduction}

In a common family of neural network language models, the current input word is represented as the vector $c\in {\rm I\!R}^{C}$ and is projected to a dense representation using a word embedding matrix $U$. Some computation is then performed on the word embedding $U^\top c$, which results in a vector of activations $h_2$. A second matrix $V$ then projects $h_2$ to a vector $h_3$ containing one score per vocabulary word: $h_3=V h_2  $. The vector of scores is then converted to a vector of probability values $p$, which represents the models' prediction of the next word, using the softmax function.

For example, in the LSTM-based language models of~\cite{sundermeyer12:lstm,zaremba14}, for vocabulary of size $C$, the one-hot encoding is used to represent the input $c$ and $U \in {\rm I\!R}^{C \times H}$. An LSTM is then employed, which results in an activation vector $h_2$ that similarly to $U^\top c$, is also in ${\rm I\!R} ^ H$. %
In this case, $U$ and $V$ are of exactly the same size.

We call $U$ the input embedding, and $V$ the output embedding. In both matrices, we expect rows that correspond to similar words to be similar: for the input embedding, we would like the network to react similarly to synonyms, while in the output embedding, we would like the scores of words that are interchangeable to be similar~\cite{mnih2012fast}.

While $U$ and $V$ can both serve as word embeddings, in the literature, only the former serves this role. In this paper, we compare the quality of the input embedding to that of the output embedding, and we show that the latter can be used to improve neural network language models. Our main results are as follows:
(i) We show that in the word2vec skip-gram model, the output embedding is only slightly inferior to the input embedding. This is shown using metrics that are commonly used in order to measure embedding quality.
(ii) In recurrent neural network based language models, the output embedding outperforms the input embedding.
(iii) By tying the two embeddings together, i.e., enforcing $U=V$, the joint embedding evolves in a more similar way to the output embedding than to the input embedding of the untied model.
(iv) Tying the input and output embeddings leads to an improvement in
the perplexity of various language models. This is true both when using dropout or when not using it.
(v) When not using dropout, we propose adding an additional projection $P$ before $V$, and apply regularization to $P$. 
(vi) Weight tying in neural translation models can reduce their size (number of parameters) to less than half of their original size without harming their performance. %

\section{Related Work}

Neural network language models (NNLMs) assign probabilities to word sequences. Their resurgence was initiated by~\cite{Bengio:2003}. Recurrent neural networks were first used for language modeling in~\cite{mikolov10inters} and~\cite{Pascanu13}. The first model that implemented language modeling with LSTMs~\cite{LSTM97} was~\cite{sundermeyer12:lstm}. Following that,~\cite{zaremba14} introduced a dropout~\cite{dropout} augmented NNLM.~\cite{yarin2,yarin1} proposed a new dropout method, which is referred to as Bayesian Dropout below, that improves on the results of~\cite{zaremba14}.

The skip-gram word2vec model introduced in~\cite{word2vec,word2vec2} learns representations of words. This model learns a representation for each word in its vocabulary, both in an input embedding matrix and in an output embedding matrix. When training is complete, the vectors that are returned are the input embeddings. The output embedding is typically ignored, although~\cite{mitra2016dual,mnih2013learning} use both the output and input embeddings of words in order to compute word similarity. Recently,~\cite{yg14} argued that the output embedding of the word2vec skip-gram model needs to be different than the input embedding.

As we show, tying the input and the output embeddings is indeed detrimental in word2vec. However, it improves performance in NNLMs.

In neural machine translation (NMT) models~\cite{kalchbrenner2013recurrent,cho2014learning,sutskever2014sequence,learn2align}, the decoder, which generates the translation of the input sentence in the target language, is a language model that is conditioned on both the previous words of the output sentence and on the source sentence. State of the art results in NMT have recently been achieved by systems that segment the source and target words into subword units~\cite{edinburghWMT}.  One such method~\cite{bpe} is based on the byte pair encoding (BPE) compression algorithm~\cite{gage94}.  BPE segments rare words into their more commonly appearing subwords. 

Weight tying was previously used in the log-bilinear model of~\cite{mnih2009scalable}, but the decision to use it was not explained, and its effect on the model's performance was not tested. 
Independently and concurrently with our work~\cite{socher16} presented an explanation for weight tying in NNLMs based on~\cite{hinton2015distilling}.

\section{Weight Tying}
\label{sec:wt}
In this work, we employ three different model categories: NNLMs, the word2vec skip-gram model, and NMT models. Weight tying is applied similarly in all models. For translation models, we also present a three-way weight tying method.

NNLM models contain an input embedding matrix, two LSTM layers ($h_1$ and $h_2$), a third hidden scores/logits layer $h_3$, and a softmax layer. 
The loss used during training is the cross entropy loss without any regularization terms. 

Following~\cite{zaremba14}, we employ two models: large and small. The large model employs dropout for regularization. The small model is not regularized. Therefore, we propose the following regularization scheme. A projection matrix $P \in {\rm I\!R}^{H\times H}$ is inserted before the output embedding, i.e., $h_3 =  V P h_2  $. The regularizing term $\lambda \|P\|_2$ is then added to the small model's loss function. In all of our experiments, $\lambda=0.15$.

Projection regularization allows us to use the same embedding (as both the input/output embedding) with some adaptation that is under regularization. It is, therefore, especially suited for WT. 

While training a vanilla {\bf untied NNLM}, at timestep $t$, with current input word sequence $i_{1:t} = [i_1,...,i_t]$ and current target output word $o_t$, the negative log likelihood loss is given by:
${\cal L}_t = -\log  p_t(o_t|i_{1:t}) $, 
where $p_t(o_t|i_{1:t}) = \frac{\exp{(V_{o_t}^\top h_2^{(t)})}}{\sum_{x=1}^{C} \exp({{V_x^\top h_2^{(t)}}})   }$, $U_k$ ($V_k$) is the $k$th row of $U$ ($V$), which corresponds to word $k$, and $h_2^{(t)}$ is the vector of activations of the topmost LSTM layer's output at time $t$. For simplicity, we assume that at each timestep $t$, $i_t\neq o_t$. Optimization of the model is performed using stochastic gradient descent. 

The update for row $k$ of the input embedding is:

\begin{small}
\begin{equation*} %
\frac{\partial {\cal L}_t}{\partial U_k} = 
\begin{cases}

(   {\sum_{x=1}^{C} p_t(x|i_{1:t}) \cdot V_x^\top}  - V_{o_t}^\top) \frac{\partial h_2^{(t)}} {\partial U_{i_t}}      

& k = i_t  \\

 0 & k \neq i_t \\
\end{cases} 
\end{equation*}
\end{small}
For the output embedding, row $k$'s update is: 
\begin{small}
\begin{equation*}
\frac{\partial {\cal L}_t}{\partial V_k} = 
\begin{cases}
( p_t(o_t|i_{1:t}) -1 ) h_2^{(t)}      & k=o_t\\
 p_t(k|i_{1:t}) \cdot h_2^{(t)}      & k \neq o_t\\                    
\end{cases}
\end{equation*}
\end{small}
Therefore, in the untied model, at every timestep, the only row that is updated in the input embedding is the row $U_{i_t}$ representing the current input word.  This means that vectors representing rare words are updated only a small number of times. The output embedding updates every row at each timestep.

In {\bf tied NNLMs}, we set $U=V=S$. The update for each row in $S$  is the sum of the updates obtained for the two roles of S as both an input and output embedding.

The update for row $k \neq i_t$ is similar to the update of row $k$ in the untied NNLM's output embedding (the only difference being that U and V are both replaced by a single matrix S). 
In this case, there is no update from the input embedding role of $S$. 

The update for row $k=i_t$, is made up of a term from the input embedding (case $k=i_t$) and a term from the output embedding (case $ k \neq o_t$). The second term grows linearly with $p_t(i_t|i_{1:t})$, which is expected to be close to zero, since words seldom appear twice in a row (the low probability in the network was also verified experimentally). The update that occurs in this case is, therefore, mostly impacted by the update from the input embedding role of $S$.

To conclude, in the tied NNLM, every row of $S$ is updated during each iteration, and for all rows except one, this update is similar to the update of the output embedding of the untied model. This implies a greater degree of similarity of the tied embedding to the untied model's output embedding than to its input embedding. 

The analysis above focuses on NNLMs for brevity. In \textbf{word2vec}, the update rules are similar, just that $h_2^{(t)}$ is replaced by the identity function. As argued by~\cite{yg14}, in this case weight tying is not appropriate, because if $p_t(i_t|i_{1:t})$ is close to zero then so is the norm of the embedding of $i_t$. This argument does not hold for NNLMs, since the LSTM layers cause a decoupling of the input and output embedddings.  

Finally, we evaluate the effect of weight tying in \textbf{neural translation models}. 
In this model:
$p_t(o_t|i_{1:t},r) = \frac {exp(V_{o_t}^\top G^{(t)})}{\sum_{x=1}^{C_t} exp(V^\top_x G^{(t)})}   $ where $r = (r_1, ..., r_N)$ is the set of words in the source sentence,  $U$ and $V$ are the input and output embeddings of the decoder and $W$ is the input embedding of the encoder (in translation models $U,V \in {\rm I\!R}^{C_t \times H}$ and $W \in {\rm I\!R}^{C_s \times H}$, where $C_s$ / $C_t$ is the size of the vocabulary of the source / target).
$G^{(t)}$ is the decoder, which receives the context vector, the embedding of the input word ($i_t$) in $U$, and its previous state at each timestep. $c_t$ is the context vector at timestep $t$,  $ c_t =\sum _{j \in r} a_{tj} h_j$, where $a_{tj}$ is the weight given to the $j$th annotation at time $t$: $a_{tj} = \frac {\exp(e_{tj})} { \sum_{k \in r} \exp(e_{ik} )     }$, and $e_{tj} = a_t(h_j)$, where $a$ is the alignment model.  
$F$ is the encoder which produces the sequence of annotations $(h_1, ..., h_N)$.

The output of the decoder is then projected to a vector  of scores using the output embedding: $l_t = V G^{(t)}$. The scores are then converted to  probability values using the softmax function.

In our weight tied translation model, we tie the input and output embeddings of the decoder.%

We observed that when preprocessing the ACL WMT 2014 EN$\rightarrow$FR\footnote{\tiny \url{http://statmt.org/wmt14/translation-task.html}} and WMT 2015 EN$\rightarrow$DE\footnote{\tiny \url{http://statmt.org/wmt15/translation-task.html}} datasets using  BPE, many of the subwords appeared in the vocabulary of both the source and the target languages. Tab.~\ref{tab:word-count} shows that up to 90\% (85\%) of BPE subwords between English and French (German) are shared.

Based on this observation, we propose three-way weight tying (TWWT), where the input embedding of the decoder, the output embedding of the decoder and the input embedding of the encoder are all tied. The single source/target vocabulary of this model is the union of both the source and target vocabularies. In this model, both in the encoder and decoder, all subwords are embedded in the same duo-lingual space.

\begin{table}[t]

\begin{small}
\begin{center}
\begin{tabular}{l|ccc}
\hline
Language  & Subwords  & Subwords  & Subwords   \\ 
 pairs & only in source & only in target &in both \\\hline
EN$\rightarrow$FR & 2K & 7K & 85K   \\ 
EN$\rightarrow$DE & 3K & 11K & 80K  \\ \hline
\end{tabular}
\end{center}
\end{small}

\caption{\small Shared BPE subwords between pairs of languages.}
\label{tab:word-count}
\end{table}

\section{Results}
\label{sec:results}

Our experiments study the quality of various embeddings, the similarity between them, and the impact of tying them on the word2vec skip-gram model,  NNLMs, and NMT models.

\subsection{Quality of Obtained Embeddings}

In order to compare the various embeddings, we pooled five embedding evaluation methods from the literature. These evaluation methods involve calculating pairwise (cosine) distances between embeddings and correlating these distances with human judgments of the strength of relationships between concepts. We use: Simlex999~\cite{simlex999}, Verb-143~\cite{verb143}, MEN~\cite{bruni2014}, Rare-Word~\cite{Luong-rare} and MTurk-771~\cite{mturk771}. 

We begin by  training both  the  tied and untied word2vec models on the text8\footnote{\tiny \url{http://mattmahoney.net/dc/textdata}} dataset, using a vocabulary consisting only of words that appear at least five times. As can be seen in Tab.~\ref{tab:w2v-table}, the output embedding is almost as good as the input embedding. As expected, the embedding of the tied model is not competitive. The situation is different when training the small NNLM model on either the Penn Treebank~\cite{ptb} or text8 datasets (for PTB, we used the same train/validation/test set split and vocabulary as~\cite{MikolovKBCK11}, while on text8 we used the split/vocabulary from~\cite{mikolovJCMR14}). These results are presented in Tab.~\ref{tab:lm-embeddings-table}. In this case, the input embedding is far inferior to the output embedding. The tied embedding is comparable to the output embedding.   %

\begin{table}[t]

\begin{small}
\begin{center}
\begin{tabular}{l|cc|c}
\hline
                                        & Input & Output  & Tied \\ \hline
Simlex999        & 0.30           & 0.29    &    0.17      \\ 
Verb-143          & 0.41            & 0.34    &     0.12    \\ 
MEN            & 0.66           & 0.61     &    0.50     \\ 
Rare-Word     & 0.34          & 0.34     &    0.23    \\ 
MTurk-771       & 0.59            & 0.54    &    0.37     \\
\hline

\end{tabular}
\end{center}
\end{small}
\caption{\small  Comparison of input and output embeddings learned by a word2vec skip-gram model. Results are also shown for the tied word2vec model. Spearman's correlation $\rho$  is reported for five word embedding evaluation benchmarks.}
\label{tab:w2v-table}
\end{table}

\begin{table}[t]

\begin{center}
\begin{small}
\begin{tabular}{l|ll|l||ll|l}

\hline
   &   \multicolumn{3}{c||}{PTB}  &        \multicolumn{3}{c}{text8}     \\ %
Embedding & In   & Out  & Tied & In   & Out  & Tied \\ \hline
Simlex999 & 0.02 & 0.13 & 0.14 & 0.17 & 0.27 & 0.28     \\ \hline
Verb143   & 0.12 & 0.37 & 0.32 & 0.20 & 0.35 & 0.42 \\ \hline
MEN       & 0.11 & 0.21 & 0.26 & 0.26 & 0.50 & 0.50    \\ \hline
Rare-Word & 0.28 & 0.38 & 0.36 & 0.14 & 0.15 & 0.17 \\ \hline
MTurk771  & 0.17 & 0.28 & 0.30 & 0.26 & 0.48 & 0.45 \\ \hline
\end{tabular}
\end{small}
\end{center}
\caption {\small Comparison of the input/output embeddings of the small model from~\protect\cite{zaremba14} and the embeddings from our weight tied variant. Spearman's correlation $\rho$ is presented.}
\label{tab:lm-embeddings-table}

\end{table}

\begin{table}[t]

\begin{center}
\begin{small}
\begin{tabular}{lc|ccc}
\hline
A                & B                     & $\rho(A,B)$ & $\rho(A,B)$ & $\rho(A,B)$ \\ 
& & word2vec & NNLM(S) & NNLM(L) \\ \hline
In   & Out      & 0.77&0.13 & 0.16                          \\ 
In   & Tied & 0.19& 0.31& 0.45                        \\ 
Out  & Tied & 0.39 &0.65 & 0.77                         \\ \hline
\end{tabular}
\end{small}
\end{center}
\caption{\small Spearman's rank correlation $\rho$ of similarity values between all pairs of words evaluated for the different embeddings: input/output embeddings (of the untied model) and the embeddings of our tied model. We show the results for both the word2vec models and the small and large NNLM models from~\protect\cite{zaremba14}.}
\label{tab:emb-dist-spearm}
\end{table}

A natural question given these results and the analysis in Sec.~\ref{sec:wt} is whether the word embedding in the weight tied NNLM model is more similar to the input embedding or to the output embedding of the original model. We, therefore, run the following experiment: First, for each embedding, we compute the cosine distances between each pair of words.  We then compute Spearman's rank correlation between these vectors of distances. As can be seen in Tab.~\ref{tab:emb-dist-spearm}, the results are consistent with our analysis and the results of Tab.~\ref{tab:w2v-table} and Tab.~\ref{tab:lm-embeddings-table}: for word2vec the input and output embeddings are similar to each other and differ from the tied embedding; for the NNLM models, the output embedding and the tied embeddings are similar, the input embedding is somewhat similar to the tied embedding, and differs considerably from the output embedding.

\subsection{Neural Network Language Models}

\begin{table}[t]

\begin{center}
\begin{small}
\tabcolsep=0.11cm

\begin{tabular}{l|cccc}
\hline
Model           &Size&                              Train          & Val. & Test \\
\hline
Large~\protect\cite{zaremba14}           &66M &  37.8       & 82.2                   & 78.4             \\ 
Large + Weight Tying      &51M &  48.5    & 77.7                  & 74.3            \\ 
\hline
Large + BD~\protect\cite{yarin2} + WD &66M &    24.3      &  78.1                 &   75.2          \\ 
Large + BD + WT &51M &    28.2    &   75.8          &   73.2         \\ 
\hline
RHN~\protect\cite{rhn} + BD        &32M & 67.4 &71.2	&68.5\\
RHN + BD + WT   &24M & 74.1 & 68.1 &	66.0\\
\hline
\end{tabular}
\end{small}
\end{center}
\caption{\small Word level perplexity (lower is better) on PTB and size (number of parameters) of models that use either dropout (baseline model) or Bayesian dropout (BD). WD -- weight decay.}
\label{tab:lm-ptb-do}
\end{table}

\begin{table}[t]

\begin{center}
\begin{small}

\begin{tabular}{l|cccc}
\hline
     Model  &   Size & Train     & Val.  & Test\\ \hline
KN 5-gram   &   &  &                       & 141             \\ 
RNN         &   &            &          & 123             \\ 
LSTM     &   &         &          & 117             \\ 
Stack RNN   &8.48M  &  &                       & 110             \\ 
FOFE-FNN    &   &   &                       & 108             \\ 
Noisy LSTM  &4.65M  &      &  111.7                & 108.0 \\ 
Deep RNN    &6.16M  &   &                 & 107.5             \\ 
Small model &4.65M  & 38.0 & 120.7                & 114.5          \\ 
\hline
Small  + WT &2.65M & 36.4&  117.5  &   112.4\\  
Small  + PR &4.69M   & 50.8 & 116.0   &  111.7  \\ 

Small  + WT + PR & 2.69M & 53.5 & 104.9   & 100.9  \\ \hline
\end{tabular}
\end{small}
\end{center}

\caption{\small Word level perplexity on PTB and size for models that do not use dropout. The compared models are: KN 5-gram~\protect\cite{MikolovKBCK11}, RNN~\protect\cite{MikolovKBCK11}, LSTM~\protect\cite{Graves13}, Stack / Deep RNN~\protect\cite{Pascanu13}, FOFE-FNN~\protect\cite{Zhang15}, Noisy LSTM~\protect\cite{gulcehre16}, and the small model from ~\protect\cite{zaremba14}. The last three models are our models, which extend the small model. PR -- projection regularization. }
\label{tab:lm-ptb-no-do}
\end{table}

\begin{table}[t]

\tabcolsep=0.11cm
\begin{center}
\begin{small}
\begin{tabular}{l|l|c|ccc}
\hline
      & Model   & Small & S + WT & S + PR & S + WT + PR \\ \hline
\parbox[t]{2mm}{\multirow{3}{*}{\rotatebox[origin=c]{90}{text8}}} 
      & Train & 90.4 &   95.6     &  92.6      &  95.3            \\ 
      & Val.  &  -  &      -   &      -    &     -          \\ 
      & Test  &  195.3  &   187.1     &   199.0 &  183.2           \\ \hline
\parbox[t]{2mm}{\multirow{3}{*}{\rotatebox[origin=c]{90}{IMDB}}}  
      & Train  &  71.3  & 75.4        &   72.0   & 72.9              \\ 
      & Val.   &  94.1  & 94.6        &  94.0      &  91.2          \\ 
      & Test   &  94.3  & 94.8      &  94.4     &  91.5         \\ \hline
\parbox[t]{2mm}{\multirow{3}{*}{\rotatebox[origin=c]{90}{BBC}}}   
      & Train  &  28.6     &    30.1        & 42.5           &    45.7     \\ 
      & Val.   & 103.6      &   99.4      &   104.9     &96.4                 \\ 
      & Test   &  110.8     & 106.8           & 108.7            &   98.9              \\ \hline
\end{tabular}
\end{small}
\end{center}
\caption{\small Word level perplexity on the text8, IMDB and BBC datasets. The last three models are our models, which extend the small model (S) of~\protect\cite{zaremba14}.} %
\label{tab:lm-text8-no-do}
\end{table}
We next study the effect of tying the embeddings on the perplexity obtained by the NNLM models. Following~\cite{zaremba14}, we study two NNLMs. The two models differ mostly in the size of the LSTM layers. In the small model, both LSTM layers contain 200 units and in the large model, both contain 1500 units. In addition, the large model uses three dropout layers, one placed right before the first LSTM layer, one between $h_1$ and $h_2$ and one right after $h_2$. The dropout probability is $0.65$. 
For both the small and large models, we use the same hyperparameters (i.e. weight initialization, learning rate schedule, batch size) as in~\cite{zaremba14}.

In addition to training our models on PTB and text8, following~\cite{Miyamoto16}, we also compare the performance of the NNLMs on the BBC~\cite{BBC} and IMDB~\cite{IMDB} datasets, each of which we process and split into a train/validation/test split (we use the same vocabularies as~\cite{Miyamoto16}).  %

In the first experiment, which was conducted on the PTB dataset, we compare the perplexity obtained by the large NNLM model and our version in which the input and output embeddings are tied. As can be seen in Tab.~\ref{tab:lm-ptb-do}, weight tying significantly reduces perplexity on both the validation set and the test set, but not on the training set. This indicates less overfitting, as expected due to the reduction in the number of parameters. Recently,~\cite{yarin1}, proposed a modified model that uses Bayesian dropout and weight decay. They obtained improved performance. When the embeddings of this model are tied, a similar amount of improvement is gained. We tried this with and without weight decay and got similar results in both cases, with slight improvement in the latter model. Finally, by replacing the LSTM with a recurrent highway network~\cite{rhn}, state of the art results are achieved when applying weight tying. The contribution of WT is also significant in this model. %

Perplexity results are often reported separately for models with and without dropout. In Tab.~\ref{tab:lm-ptb-no-do}, we report the results of the small NNLM model, that does not utilize dropout, on PTB. As can be seen, both WT and projection regularization (PR) improve the results. When combining both methods together, state of the art results are obtained. An analog table for text8, IMDB and BBC is Tab.~\ref{tab:lm-text8-no-do}, which shows a significant reduction in perplexity across these datasets when both PR and WT are used. PR does not help the large models, which employ dropout for regularization.

\subsection{Neural Machine Translation}
Finally, we study the impact of weight tying in attention based NMT models, using the DL4MT\footnote{\tiny \url{https://github.com/nyu-dl/dl4mt-tutorial}} implementation. %
We train our EN$\rightarrow$FR models on the parallel corpora provided by ACL WMT 2014. We use the data as processed by~\cite{cho2014learning} using the data selection method of~\cite{axelrod11}. For EN$\rightarrow$DE we train on data from the translation task of WMT 2015, validate on newstest2013 and test on newstest2014 and newstest2015. Following~\cite{bpe} we learn the BPE segmentation on the union of the vocabularies that we are translating from and to (we use BPE with 89500 merge operations). %
All models were trained using Adadelta~\cite{adadelta} for 300K updates, have a hidden layer size of 1000 and all embedding layers are of size 500.

\begin{table}[t]
\begin{small}
\begin{center}
\begin{tabular}{ll|ccc}
\hline
      &          & Size & Validation & Test  \\ \hline
EN$\rightarrow$FR & Baseline & 168M  & 29.49 & 33.13     \\ 
                  & Decoder WT        & 122M  & 29.47 & 33.26   \\ 
                  & TWWT     & 80M   & 29.43 & 33.46 \\ \hline

EN$\rightarrow$DE & Baseline & 165M  & 20.96 & 16.79    \\ 
                  & Decoder WT       & 119M  & 21.09 & 16.54   \\ 
                  & TWWT     & 79M   & 21.02 & 17.15 \\ \hline
\end{tabular}
\end{center}
\end{small}
\caption{\small Size (number of parameters) and BLEU score of various translation models. TWWT -- three-way weight tying.}

\label{tab:translation}
\end{table}

Tab.~\ref{tab:translation} shows that even though the weight tied models have about 28\% fewer parameters than the baseline models, their performance is similar. This is also the case for the three-way weight tied models, even though they have about 52\% fewer parameters than their untied counterparts.

\bibliography{eacl2017}
\bibliographystyle{eacl2017}

\end{document}